\documentclass[10pt,twocolumn,letterpaper]{article}

\usepackage{cvpr}              %

\usepackage{multirow}
\usepackage{colortbl}
\usepackage{siunitx}
\usepackage{pifont}
\usepackage{arydshln} %
\usepackage{bm}
\usepackage{wrapfig}

\definecolor{first}{rgb}{1, 0.7, 0.7}
\definecolor{second}{rgb}{1,0.85, 0.7}
\definecolor{third}{rgb}{1,1, 0.8}

\newcommand{\token}{{\mathbf{T}}}

\newif\ifshowcomments
\showcommentstrue

\newcommand{\parahead}[1]{\vspace{1mm}\noindent\textbf{{#1}.}\ }

\newcommand{\MethodName}{TokenGS\xspace}

\definecolor{cvprblue}{rgb}{0.21,0.49,0.74}
\usepackage[pagebackref,breaklinks,colorlinks,allcolors=cvprblue]{hyperref}
\usepackage[accsupp]{axessibility}  %

\def\paperID{5994} %
\def\confName{CVPR}
\def\confYear{2026}

\title{\resizebox{\textwidth}{!}{\MethodName: Decoupling 3D Gaussian Prediction from Pixels with Learnable Tokens}}

\author{
    Jiawei Ren\thanks{Equal contribution.} \quad
    Michal Jan Tyszkiewicz\footnotemark[1] \quad
    Jiahui Huang\thanks{Equal advising.} \quad
    Zan Gojcic\footnotemark[2] \\
    NVIDIA \\
    {\tt\small \{jiaweir, mtyszkiewicz, jiahuih, zgojcic\}@nvidia.com} \\
    { Project: \href{https://research.nvidia.com/labs/toronto-ai/tokengs}{\texttt{research.nvidia.com/labs/toronto-ai/tokengs}}}
}

\begin{document}

\twocolumn[{%
\renewcommand\twocolumn[1][]{#1}%

\maketitle
\vspace{-2.2em}
\centering
  \includegraphics[width=\textwidth]{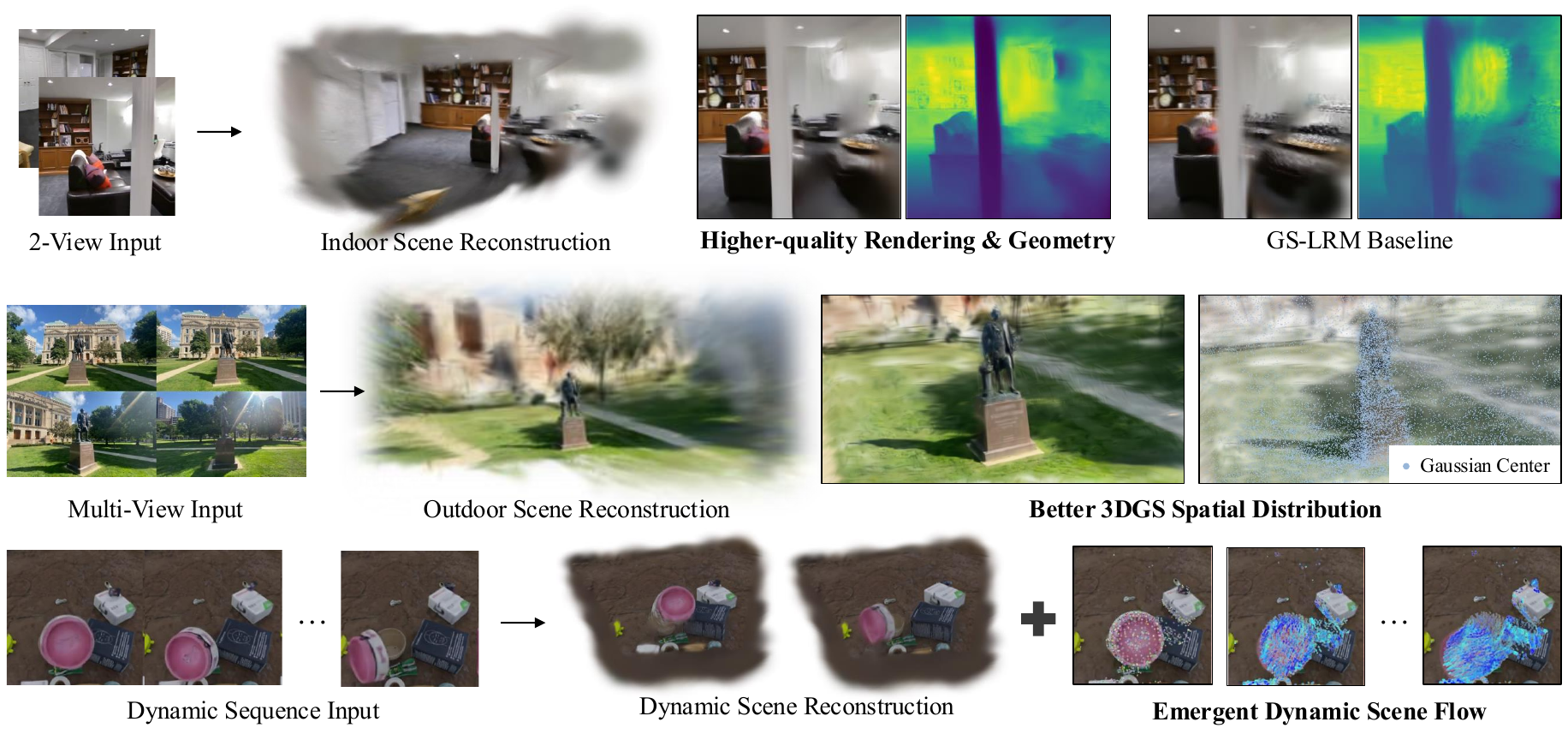}
  \vspace{-1.8em}
  \captionsetup{type=figure}
  \captionof{figure}{
  \textbf{\MethodName} is a feed-forward reconstruction framework that outputs a 3D Gaussian Splatting (3DGS) representation from posed input images. Our novel encoder-decoder architecture detaches 3D Gaussians from input pixels and enables multiple properties desirable for 3D reconstruction, as demonstrated in each example.
  }
  \label{fig:teaser}
 \vspace{1.8em}
 }
]

{
\renewcommand{\thefootnote}{}%
\footnotetext{
\textsuperscript{*}Equal contribution, \textsuperscript{$\dagger$}Equal advising.
}
}

\begin{abstract}
In this work, we revisit several key design choices of modern Transformer-based approaches for feed-forward 3D Gaussian Splatting (3DGS) prediction.
We argue that the common practice of regressing Gaussian means as depths along camera rays is suboptimal, and instead propose to directly regress 3D mean coordinates using only a self-supervised rendering loss.
This formulation allows us to move from the standard encoder-only design to an encoder-decoder architecture with learnable Gaussian tokens, thereby \emph{unbinding} the number of predicted primitives from input image resolution and number of views. 
Our resulting method, \emph{\MethodName}, demonstrates improved robustness to pose noise and multiview inconsistencies, while naturally supporting efficient test-time optimization in token space without degrading learned priors. \MethodName achieves state-of-the-art feed-forward reconstruction performance on both static and dynamic scenes, producing more regularized geometry and more balanced 3DGS distribution, while seamlessly recovering emergent scene attributes such as static-dynamic decomposition and scene flow.
\end{abstract}

\section{Introduction}
\label{sec:intro}

The field of feed-forward neural reconstruction has recently seen great progress in terms of reconstruction quality~\cite{xu2024grm,zhang2025gs}, scalability to large datasets~\cite{ziwen2025long}, and support for dynamic scenes~\cite{liang2024feed,yang2024storm,lin2025movies}. 
In certain scenarios, these methods are even starting to approach the quality of computationally intensive per-scene optimization methods. 

Despite this rapid progress, the dominant paradigm of using a large encoder-only\footnote{We adopt the terminology common in Large Language Models (LLMs), referring to a sequence of self-attention layers as in \cite{dosovitskiy2020image}.} Transformer backbone to predict pixel-aligned 3D Gaussian primitives, still faces several fundamental limitations. First, \emph{predicting Gaussian means as depths along camera rays} restricts the model’s ability to internally correct for noisy camera poses and multiview-inconsistencies, and poses challenges for dynamic scenes, where points/pixels must be warped over time.
Second, the common strategy of \emph{tying the number of predicted Gaussians to the resolution and number of images} (e.g., one particle per pixel or patch) leads to an excessive number of primitives, e.g. 32 images at $512 \times 512$ resolution yield over 8 million particles. This coupling makes the representation scale with the number of input views regardless of the intrinsic scene complexity. Indeed, repeating each input $N$ times will result in $N$ times more particles for the same scene, introducing high redundancy. Finally, while \emph{feed-forward neural reconstruction naturally enables self-supervised test-time refinement, directly optimizing the Gaussian parameters tends to degrade the learned priors of the network}, particularly in low-view regime.

To address these challenges, we present \MethodName that revisits this standard formulation and incorporates several key modifications. 
Instead of predicting Gaussian means/centers as depths along the camera rays, we directly regress their 3D coordinates. This is akin to point map prediction in feed-forward Structure-from-Motion (SfM) models such as VGGT~\cite{wang2025vggt}, but importantly trained without explicit point-map ground-truth supervision, only through self-supervised rendering loss.  This formulation enables us to replace the commonly-used encoder-only backbone with a novel \emph{encoder–decoder} architecture, where a set of learnable 3DGS tokens cross-attend to compact image features and predict the corresponding 3DGS parameters. 
As a result, the number of predicted Gaussians becomes a hyperparameter independent from the input resolution and number of images. 
By decoupling the Gaussian prediction from the input images, \MethodName can freely assign Gaussians to places with higher scene complexity.
In the meantime, this unlocks scene completion beyond input views, increases robustness to noisy camera inputs, and naturally extends to dynamic content.
Moreover, the token-based formulation supports self-supervised test-time optimization by refining only the token embeddings, thereby preserving the global priors encoded by the model.
Through extensive experiments, we demonstrate that  \MethodName achieves state-of-the-art performance in both reconstruction quality and efficiency, producing compact representations with more regularized geometry, while also recovering emergent scene attributes such as scene flow.

\section{Related works}

\parahead{3D Scene Representation}%
The choice of 3D scene representation is highly coupled with the reconstruction algorithm.
Per-scene optimization methods typically adopt neural radiance fields (NeRF)~\cite{mildenhall2021nerf, barron2023zip} based on tri-plane~\cite{fridovich2023k} or hash-grid encodings~\cite{muller2022instant}, or explicit point-based representations such as 3DGS~\cite{kerbl20233d} and its variants~\cite{loccoz20243dgrt, huang20242d, yu2024mip, wu20253dgut}.
To handle dynamic scenes, these approaches extend the representation by introducing either a direct 4D extension~\cite{park2021hypernerf,duan20244d}, deformation fields from canonical geometry~\cite{wu20244d,luiten2023dynamic}, or keyframe-based interpolations~\cite{xu2024representing}.
While these representations can all achieve high-fidelity reconstruction, they are typically optimized from scratch using image supervision and optionally learned optimizers~\cite{chen2025g3r, xu2025resplat}, leading to slow reconstruction times.

\parahead{Feed-forward 3D Reconstruction}%
Feed-forward reconstruction methods infer 3D scenes directly from input images in a single forward pass. They are typically trained on large multiview datasets, enabling them to generalize to novel scenes at inference time. Early approaches focused on static scenes and were based on NeRF~\cite{yu2021pixelnerf, hong2023lrm} or more recently Gaussian splatting ~\cite{charatan2024pixelsplat, chen2025mvsplat, xu2025depthsplat, zhang2025gs, xu2024grm, jin2024lvsm, ziwen2025long, xu2025resplat}. This paradigm has also been extended to dynamic and long-horizon reconstruction, predicting spatiotemporal Gaussians or recurrently updating scene states over time~\cite{ren2024l4gm, yang2024storm, xu2025resplat}. Feed-forward design significantly reduces reconstruction latency, yet current formulations still lag behind optimization-based methods in visual fidelity, motivating the improvements proposed herein.

\parahead{Test-Time Scaling}%
Test-Time Scaling (TTS) refers to allocating additional compute at inference time to improve model performance and robustness under unseen or challenging data distributions. This concept has been most thoroughly explored in large language models~\cite{yao2023tree,wang2022self,liu2025can} and further advanced through deliberate reasoning approaches~\cite{guo2025deepseek,jaech2024openai}.
In computer vision, TTS commonly appears in form of test-time adaptation or refinement, improving both discriminative tasks~\cite{wang2020tent,sun2020test} and generative models~\cite{dalal2025one,liu2025video}.
Representative strategies include prompt tuning~\cite{jia2022visual}, slow–fast weight updates~\cite{zhang2025test}, and self-consistency augmentations~\cite{yuan2025test3r}.
For 3D reconstruction, prior methods largely rely on per-scene optimization during inference~\cite{kerbl20233d}. More recent work applies TTS to geometry foundation models through online finetuning~\cite{chen2025ttt3r,lu2024lora3d}. However, applying test-time scaling to large reconstruction models, especially those predicting 3DGS representations, remains largely unexplored.

\section{Method}
\label{sec:method}

\begin{figure*}[h]
    \centering
    \includegraphics[width=\linewidth]{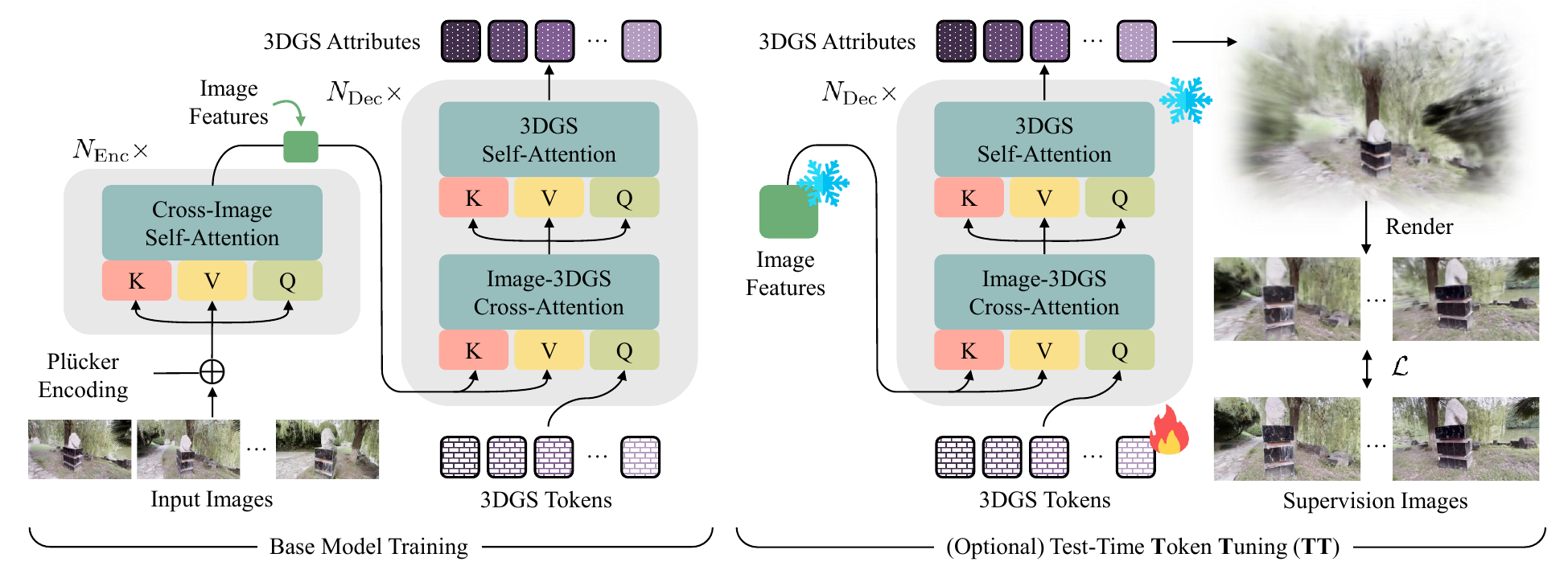}
    \vspace{-2em}
    \caption{\textbf{Our method}. We design a novel network architecture that reconstructs 3DGS from input images, directly predicting 3D Gaussian mean coordinates. The model follows an encoder-decoder structure. In the decoder, 3DGS tokens are fed in as queries to obtain the final Gaussian attributes. After the base model is trained, we allow test-time token tuning from input images to improve reconstruction quality.}
    \label{fig:architecture}
    \vspace{-1em}
\end{figure*}

Given a set of input views $\mathcal{I} = \{ \mathbf{I}_i \in \mathbb{R}^{H \times W \times 3}\}_{i=1}^N$ together with the corresponding camera extrinsics $\mathcal{T} = \{\mathbf{T}_i \in \mathbb{SE}(3)\}_{i=1}^N$ and intrinsics, our goal is to directly predict a set of $M$ 3DGS particles $\mathcal{G} \in \mathbb{R}^{M \times 14}$ from which novel views can be synthesized through volume rendering~\cite{mildenhall2021nerf, kerbl20233d}. Each Gaussian particle is parameterized by its mean $\bm{\mu} \in \mathbb{R}^3$, color $\bm{c} \in [0, 1]^3$, scale $\bm{s} \in \mathbb{R}_+^3$, opacity $\sigma \in [0, 1]$, and rotation represented as a unit quaternion $\bm{q} \in \mathbb{H}$.

We design our feed-forward 3D reconstruction framework around the idea of \emph{learnable Gaussian tokens}, thereby decoupling the positions of the Gaussian particles from the camera rays and their number from the resolution and number of input images. Specifically, we directly regress 3D Gaussian means $\bm{\mu}$ in a canonical coordinate space and train them end-to-end using volume rendering supervision and a visibility-aware regularization (\cref{sec:gaussian_mean_regression}). This formulation enables us to adopt an \textbf{encoder--decoder} architecture in which learnable Gaussian tokens cross-attend to the multi-view image tokens (\cref{sec:decoding_gaussians}). 
At inference time, our model design further allows efficient test-time tuning of the Gaussian token embeddings, improving reconstruction quality while preserving the learned priors (\cref{sec:test_time_training}). An overview of our framework is illustrated in Fig.~\ref{fig:architecture}.

\subsection{Directly Regressing Gaussian Means}
\label{sec:gaussian_mean_regression}
\begin{figure}[!t]
    \centering
    \includegraphics[width=\columnwidth]{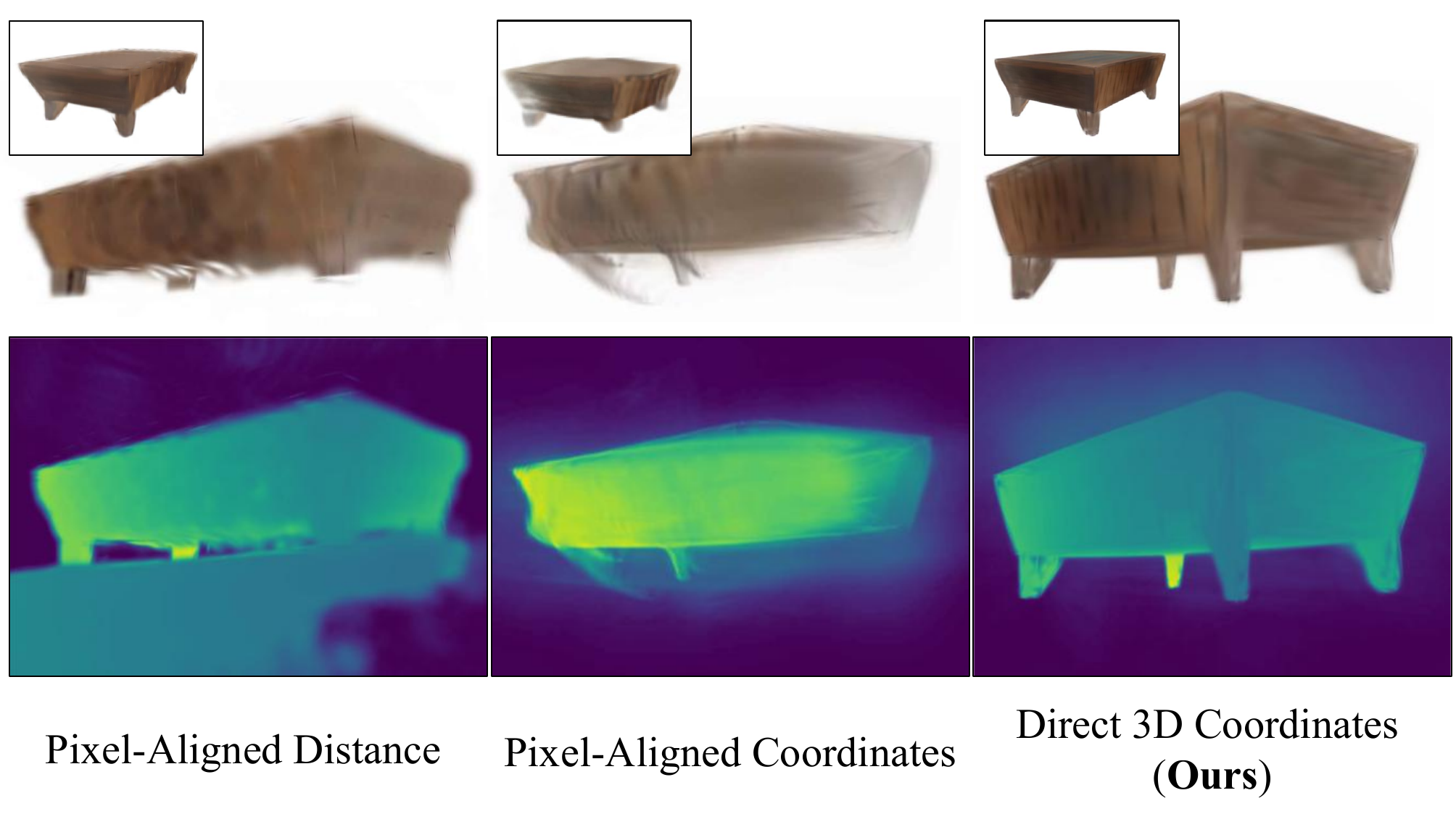}
    \vspace{-2em}
    \caption{\textbf{Comparison of 3DGS parametrizations} in feed-forward networks. While all methods reconstruct the single input view well (shown in the inset), the quality of the occluded region behind the table varies.}
    \label{fig:objaverse}
    \vspace{-2em}
\end{figure}

Instead of predicting depth along camera rays (or depth plus 3D offsets), we directly regress the 3D coordinates of Gaussian means $\bm{\mu}$ in a global coordinate frame defined by the camera extrinsics and shared across all views. 
To demonstrate its effectiveness, we did a preliminary study on Objaverse~\cite{deitke2023objaverse} dataset, where three feed-forward models using different output parametrizations are trained and compared.
As shown in \cref{fig:objaverse}, with only one view as input, the representation power of pixel-aligned parametrization is limited, not being able to recover the occluded part of the geometry.

Decoupling the Gaussian mean prediction from the camera rays offers several additional benefits: (i) it enables extrapolation and scene completion (\cf \cref{fig:view-extrapolation}), (ii) improves robustness to noise in the camera poses (\cf \cref{fig:camera-noise}), and (iii) eliminates the spiky artifacts commonly observed in depth-prediction networks (\cf \cref{fig:re10k-qual}).

\paragraph{Zero-Gradient Problem.} 
Despite the above advantages, training a direct coordinate regression formulation is non-trivial. Gaussians that lie outside all camera frusta do not contribute to the rendered supervision images and therefore receive zero gradients from the rendering loss. Such ``inactive'' Gaussians degrade training stability, waste model capacity, and manifest as floating, noisy points around the scene. 

Recent feed-forward SfM approaches~\cite{wang2025vggt} mitigate this issue by supervising their point-map outputs with explicit 3D supervision obtained by transforming depth map labels into the global coordinate frame. However, such supervision is difficult to obtain in real-world settings (or it is noisy) and reintroduces the bias of forcing the predicted means to lie along camera rays. This is precisely the coupling that our formulation seeks to avoid.\footnote{In the token formulation introduced in \cref{sec:decoding_gaussians}, we decouple 3D points from pixels, removing their correspondence and necessitating a Chamfer or similar distance for supervision, which adds further complexity.}

\paragraph{Visibility Loss.}
Instead, we introduce a \emph{visibility loss} that softly constrains Gaussian particles to remain visible in at least one of the supervision views. Specifically, we project the centroids $\bm{\mu}_m = (x_m, y_m, z_m)$ of the Gaussians to image plane of all supervision views $\mathbf{I}_i \in \mathcal{I}_\text{sup}$ to obtain their image coordinates $(u_m^i, v_m^i)$.
After normalization by image width $W$ and height $H$:
\begin{equation}
    \tilde{u}_m^i = 2(u_m^i / W) - 1, \qquad
    \tilde{v}_m^i = 2(v_m^i / H) - 1,
\end{equation}
points inside the image plane satisfy $\tilde{u}, \tilde{v} \in [-1, 1]$.
The visibility loss measures the minimum distance from each Gaussian to the nearest visible boundary:
\begin{equation}
    \mathcal{L}_{\text{vis}} =
    \sum_m \min_{\mathbf{I}_i \in \mathcal{I}_{sup}}
    \left[ \mathrm{ReLU}(|\tilde{u}_m^i| - 1)
    + \mathrm{ReLU}(|\tilde{v}_m^i| - 1) \right].
\end{equation}
\begin{wrapfigure}{r}{0.5\linewidth}
\vspace{-1em}
\hspace*{-.75\columnsep}\includegraphics[width=1.2\linewidth]{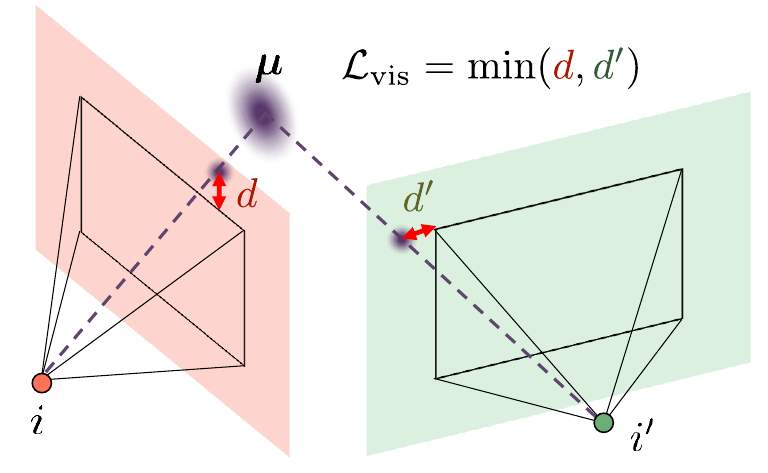}
\vspace{-1em}
\end{wrapfigure}
This penalty is zero when the Gaussian projects inside at least one view and provides gradients to pull it back into view otherwise.
In practice, to avoid spurious gradients, we clip the loss value with a constant factor $1.0$.
A conceptual illustration is shown in the inset figure and the effectiveness is demonstrated with \cref{fig:visibility-ablation}.

Besides the novel visibility loss, we supervise on renderings with a combination of pixel-wise mean squared error and SSIM losses:
\(
    \mathcal{L} = \mathcal{L}_\text{MSE} + \lambda_\text{SSIM} \mathcal{L}_\text{SSIM} + \lambda_\text{vis} \mathcal{L}_\text{vis}.\)

\subsection{Decoding Learnable Gaussian Tokens}
\label{sec:decoding_gaussians}
We employ an \textbf{encoder--decoder Transformer} where the encoder converts input images and camera parameters into image tokens, and the decoder predicts a set of learnable Gaussian tokens through cross-attention. Each Gaussian token decodes independently of the number of input pixels, effectively decoupling the number of output Gaussians from image resolution.

\subsubsection{Model Architecture}
We use a ViT-based encoder~\cite{dosovitskiy2020image}.
Specifically, each input view $\{\mathbf{I}_v \in \mathcal{I}_c\}_{v=1}^{N_c}$ is first patchified into a set of patches $\{\mathbf{I}_{vk} \in \mathbb{R}^{p \times p \times 3}\}_{k=1}^{HW/p^2}$, for a total of $N_\mathbf{I} = N_cHW/p^2$ patches.
These are projected into a $C$-dimensional latent space using a linear layer $\mathbf{x}_{vk} = \mathrm{Linear}(\mathbf{I_{vk}})$.
In parallel, we patchify the Pl\"ucker coordinates associated with each view into a set of patches  $\{\mathbf{P}_{vk} \in \mathbb{R}^{p \times p \times 6}\}_{k=1}^{HW/p^2}$ and project them as $\mathbf{s}_{vk} = \mathrm{Linear}(\mathbf{P_{vk}})$.
The two embeddings are then summed $\mathbf{a}_{vk} = \mathbf{x}_{vk} + \mathbf{s}_{vk}$ and flattened into a 1D sequence of encoder input tokens $\mathbf{A} \in \mathbb{R}^{N_\mathbf{I} \times C}$. We apply standard ViT layers, allowing attention across all views, and finally apply layer normalization to obtain the \emph{image tokens} $\mathbf{B} \in \mathbb{R}^{N_\mathbf{I} \times C}$.

Our decoder follows the DETR design~\cite{carion2020end}. We initialize $N_t$ learnable token embeddings $\mathbf{T_{\text{IN}}} \in \mathbb{R}^{N_t \times C}$ (which we refer to as \textit{3DGS tokens}) and apply Transformer decoder blocks~\cite{vaswani2017attention} consisting of (i) cross-attention to the image tokens $\mathbf{B}$, (ii) self-attention among the GS tokens, and (iii) per-token MLP.
This produces output embeddings $\mathbf{T_{\text{OUT}}}$.

From each output embedding, we regress the 14 Gaussian attributes for $N_\mathcal{G}=64$ Gaussian primitives using a linear layer. As these are real-valued, we need to map each attribute to its domain.
Following \cite{wang2025vggt}, we predict XYZ coordinates using $f(x) = \text{sign}(x) \cdot (\exp(x)-1)$. For the remaining 3DGS attributes we follow~\cite{zhang2025gs}: scaled \texttt{tanh} for color $\bm{c}$ and opacity $\bm{\sigma}$, clipped exponential for the scale $\bm{s}$, and unit normalization for the rotation quaternion $\bm{q}$. After this mapping, we flatten the Gaussians from all tokens to obtain $\mathcal{G} \in \mathbb{R}^{N_c N_g \times 14}$, introduced at the beginning of ~\cref{sec:method}.

Because there are typically far more image tokens than GS tokens ($N_\mathbf{I} \gg N_t$), making decoder's memory footprint depend primarily on $N_t$ allows for scaling to deeper architecture.
Thus, we use an optimization in which all decoder cross-attention layers share their key–value projections of image tokens.
These projections are computed once at the start of decoding, which for a decoder of depth $D_\mathrm{dec}$ replaces an allocation of size $O(N_\mathbf{I} D_\mathrm{dec})$ with a single $O(N_\mathbf{I})$.

For both the encoder and the decoder, we use LayerScale~\cite{touvron2021going} and QK-normalization~\cite{henry2020query}, which we find crucial for stable training.
Contrary to standard practice, we omit the layer-normalization before the final regression head, as we find it degrades reconstruction quality.

\subsubsection{Dynamic Scene Modeling}
\begin{figure}[t]
    \centering
    \includegraphics[width=0.9\columnwidth]{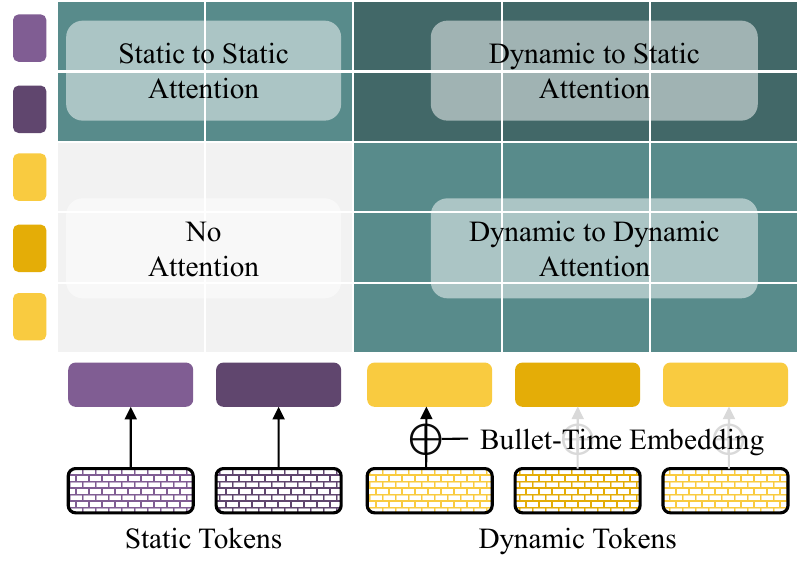}
    \caption{\textbf{Attention masking} for dynamic scenes. The figure shows one instantiation of the 3DGS self-attention block. The horizontal axis shows queries, while the vertical axis shows keys/values of the corresponding tokens.}
    \label{fig:dynamic-attn}
    \vspace{-1em}
\end{figure}

For dynamic scenes, we extend the framework with \emph{time-conditioned dynamic 3DGS tokens} following a bullet-time reconstruction formulation~\cite{liang2024feed}.
Each dynamic token receives a learnable time embedding representing the target frame, while the static tokens remain time-invariant.
Dynamic tokens attend causally to static ones, allowing the model to decompose the scene into static and motion components while maintaining consistent correspondences across time.
A schematic figure is shown in \cref{fig:dynamic-attn}.

\parahead{Static and Dynamic Tokens}
Concretely, we split the original set of 3DGS tokens $\token$ into two separate sets of tokens:
\begin{equation}
\begin{aligned}
    \token^{\mathrm{S}} = \{\, \token^{\mathrm{S}}_1,\ldots,\token^{\mathrm{S}}_{N_s} \,\} \in \mathbb{R}^{N_s \times C}, \\
    \token^{\mathrm{D}} = \{\, \token^{\mathrm{D}}_1,\ldots,\token^{\mathrm{D}}_{N_d}\,\} \in \mathbb{R}^{N_d \times C},
\end{aligned}
\end{equation}
where $\token^{\mathrm{S}}$ is intended to represent the \emph{static} (time-invariant) part of the scene and $T^{\mathrm{D}}$ represents the dynamic part.
We borrow the formulation introduced in \cite{liang2024feed}, and augment the dynamic token with the timestamp $t$ (\ie \emph{bullet-time}) that one wants to reconstruct.
Specifically, we add a learned time embedding as follows:
\begin{equation}
    \tilde \token^{\mathrm{D}}_j(t)
   = \token^{\mathrm{D}}_j + \mathrm{Linear}(\tau(t)),
\end{equation}
where $\tau(t) \in \mathbb{R}^{d_\tau}$ represents sinusoidal encoding which gets linearly projected to the latent space.

\parahead{Attention Masking}
We impose a structured attention mask over the combined token set $\token(t) = \token^{\mathrm{S}} \cup \tilde \token^{\mathrm{D}}(t)$ in the self-attention part of our Transformer-decoder. 
Specifically, we impose a causal structure where the dynamic tokens can only attend/query unidirectionally to the static tokens.
This provides inductive bias that the dynamic part of the scene is dependent on the static part, enabling the model to decompose a scene into time-invariant
(static) structure and time-varying (dynamic) motion. The dynamic tokens preserve consistent correspondences across all frames.

\subsection{Test-Time Scaling}
\label{sec:test_time_training}
Our encoder-decoder design naturally supports two complementary forms of test-time scaling without requiring network retraining. The first, \textbf{context extension}, is achieved by supplying more image tokens at inference time than were used during training (e.g., providing 4 views instead of 2). In alternative formulations (e.g,  encoder-only design) this would increase the number of predicted Gaussians, whereas in our formulation the number of Gaussians remains fixed and independent of the context length. The second, \textbf{token-tuning (TT)}, is a lightweight test-time training procedure in which we fine-tune only the Gaussian token embeddings using self-supervision on the input views, with the network parameters and image features frozen. A small number of gradient steps adapts the attention patterns to the specific scene, leading to improved Gaussian parameters while preserving the learned priors encoded in the rest of the network. 
Together, these two strategies provide flexible test-time scaling and yield substantial improvements in reconstruction quality, with fidelity increasing as more input views become available.

\section{Experiments}

\begin{table}[t]
\caption{\textbf{Reconstruction performance with two input views on RealEstate10K~\cite{zhou2018stereo}.} 
Resolution is $256 \times 256$.}
\vspace{-1em}
\centering
\scriptsize
\setlength{\tabcolsep}{8pt}
\begin{tabular}{lccc c}
\toprule
Method & PSNR$\uparrow$ & SSIM$\uparrow$ & LPIPS$\downarrow$ & \#GS \\
\midrule
MVSplat~\citep{chen2025mvsplat} & 26.39 & 0.869 & 0.128 & 131K\\
DepthSplat~\citep{xu2025depthsplat} & 27.47 & 0.889 & \textbf{0.114} & 131K\\
GS-LRM~\citep{zhang2025gs} & 28.10 & 0.892 & \textbf{0.114} & 131K\\
\midrule
\textbf{Ours} (1024 tok) & 28.02 & 0.896 & 0.147 & 66K\\
\textbf{Ours} (4096 tok) & \textbf{28.41} & \textbf{0.903} & 0.135 & 262K\\
\hdashline
\textbf{Ours} (4096 tok, +TT) & 28.82 & 0.910 & 0.130 & 262K\\
\bottomrule
\end{tabular}
\label{tab:sota_re10k}
\vspace{-1em}
\end{table}

\begin{table}[t]

    \begin{center}
    \small
    \caption{
    \textbf{Evaluations on DL3DV~\cite{ling2024dl3dv} dataset with different number of input views.}
    \textbf{Ours} is trained on 4-view, 2- and 6-view experiments show generalization to other context lengths (\textbf{Ours*}).
    Prior work uses separate models for each \#Views.
    Resolution is $448 \times 256$.
    }
    \vspace{-1em}
    \resizebox{\linewidth}{!}{
    \begin{tabular}{lcccccccccccccccccccccccc}
    \toprule
    Method & \#Views & PSNR $\uparrow$ & SSIM $\uparrow$ & LPIPS $\downarrow$ & Time (s) & \#GS \\
        
    \toprule
    
    MVSplat~\cite{chen2025mvsplat} & \multirow{4}{*}[-2pt]{2} & 17.54 & 0.529 & 0.402 & 0.072 & 229k \\
    DepthSplat~\cite{xu2025depthsplat} & & 19.31 & \textbf{0.615} & \textbf{0.310} & 0.083 & 229k \\
    \textbf{Ours}$^*$ & & \textbf{19.35} & 0.609 & 0.440 & \textbf{0.049} & 262k \\
    \hdashline
    \textbf{Ours}$^*$ (+TT) & & 19.32 & 0.601 & 0.435 & 4.00 & 262k \\
    
    \midrule
    
    MVSplat~\cite{chen2025mvsplat} & \multirow{4}{*}[-2pt]{4} & 21.63 & 0.721 & 0.233 & 0.146 & 458k \\
    DepthSplat~\cite{xu2025depthsplat} & & \textbf{23.12} & \textbf{0.780} & \textbf{0.178} & 0.107 & 458k \\
    \textbf{Ours} & & 23.02 & 0.747 & 0.326 & \textbf{0.075} & 262k \\
    \hdashline
    \textbf{Ours} (+TT) & & 23.57 & 0.766 & 0.314 & 4.82 & 262k \\
    
    \midrule
    
    MVSplat~\cite{chen2025mvsplat} & \multirow{4}{*}[-2pt]{6} & 22.93 & 0.775 & 0.193 & 0.263 & 688k \\
    DepthSplat~\cite{xu2025depthsplat} & & \textbf{24.19} & \textbf{0.823} & \textbf{0.147} & 0.132 & 688k \\
    \textbf{Ours}$^*$ & & 23.69 & 0.766 & 0.311 & \textbf{0.085} & 262k \\
    \hdashline
    \textbf{Ours}$^*$ (+TT) & & 24.51 & 0.792 & 0.295 & 5.61 & 262k \\

    \bottomrule
    \end{tabular}
    \label{tab:dl3dv_compare}
    
    }
    \end{center}
    \vspace{-1em}
\end{table}

\begin{table}[t]
    \caption{\textbf{View extrapolation evaluation with two input views on DL3DV and RealEstate10K.} Ours outperforms models finetuned with PM-Loss~\cite{shi2025revisiting}, a point map supervision loss.}
    \vspace{-1em}
    \centering
    \footnotesize
    \setlength{\tabcolsep}{2.5pt} 
    \begin{tabular}{l ccc c ccc}
    \toprule
    \multirow{2}{*}[-2pt]{{Method}} & \multicolumn{3}{c}{\textbf{DL3DV~\cite{ling2024dl3dv}}} && \multicolumn{3}{c}{\textbf{RealEstate10K~\cite{zhou2018stereo}}} \\
    \addlinespace[-12pt] \\
    \cmidrule{2-4} \cmidrule{6-8}
    \addlinespace[-12pt] \\
    & PSNR$\uparrow$ & SSIM$\uparrow$ & LPIPS$\downarrow$ && PSNR$\uparrow$ & SSIM$\uparrow$ & LPIPS$\downarrow$ \\ \midrule
    DepthSplat & 18.46 & 0.689 & 0.261 && 20.43 & 0.788 & 0.218 \\
    MVSplat & 16.79 & 0.592 & 0.322 && 19.52 & 0.757 & 0.231 \\
    MVSplat+PM & 19.25 & 0.615 & 0.291 && 22.18 & 0.787 & 0.199 \\
    DepthSplat+PM & 20.77 & \textbf{0.705} & \textbf{0.245} && 22.48 & 0.814 & \textbf{0.194} \\
    \midrule
    \textbf{Ours} & \textbf{20.98} & 0.688 & 0.364 && \textbf{22.60} & 0.831 & 0.222 \\
    \bottomrule
    \end{tabular}
    \label{tab:extrapolation}
    \vspace{-1em}
\end{table}

\begin{table}[t]
\caption{\textbf{Dynamic reconstruction on 4-view Kubric 4D~\citep{liang2024feed}}.}
\vspace{-1em}
\centering
\footnotesize
\setlength{\tabcolsep}{8pt}
\begin{tabular}{lccc}
\toprule
Method & PSNR$\uparrow$ & SSIM$\uparrow$ & LPIPS$\downarrow$ \\
\midrule
BTimer~\citep{liang2024feed} & 24.45 & \textbf{0.835} & \textbf{0.162} \\
\midrule
\textbf{Ours} & \textbf{24.84} & 0.821 & 0.287 \\
\bottomrule
\end{tabular}
\label{tab:kubric}
\vspace{-1em}
\end{table}

\parahead{Implementation details}%
Our model has 6 encoder layers and 24 decoder layers with channel dimension $C=1024$. To test with higher resolution and more input views on DL3DV~\cite{ling2024dl3dv}, we reduce the encoder and decoder depth to 3 and 12, respectively. We apply a z-axis offset to the XYZ activation function to initialize Gaussians in front of the reference camera whose pose is set to identity.
We use FlashAttention2~\cite{dao2023flashattention} and FlexAttention~\cite{he2024flexattention} for attention and masked attention computation.

\parahead{Training details}%
We optimize the model with AdamW~\cite{loshchilov2017decoupled} optimizer.
We use a batch size of 64 across 8 NVIDIA A100 GPUs. We first train our model for 150K iterations with 1024 Gaussian tokens on resolution $256 \times 256$, and then finetune it for 10K iterations with 4096 Gaussian tokens. We change the resolution to $448 \times 256$ in finetuning for DL3DV to match the baselines. We use a cosine learning rate scheduler with an initial learning rate of $4\cdot 10^{-4}$ and 2,000 warmup iterations. We reduce the learning rate and warmup iterations to $4 \cdot 10^{-5}$ and 400, respectively, for finetuning. For token tuning, we use 50 steps and lr=$10^{-4}$ unless indicated otherwise. We use $\lambda_\text{SSIM}=0.2$ and $\lambda_\text{vis}=1.0$.

\subsection{Static Scene Reconstruction}
\begin{figure}[!t]
    \centering
    \includegraphics[width=\linewidth]{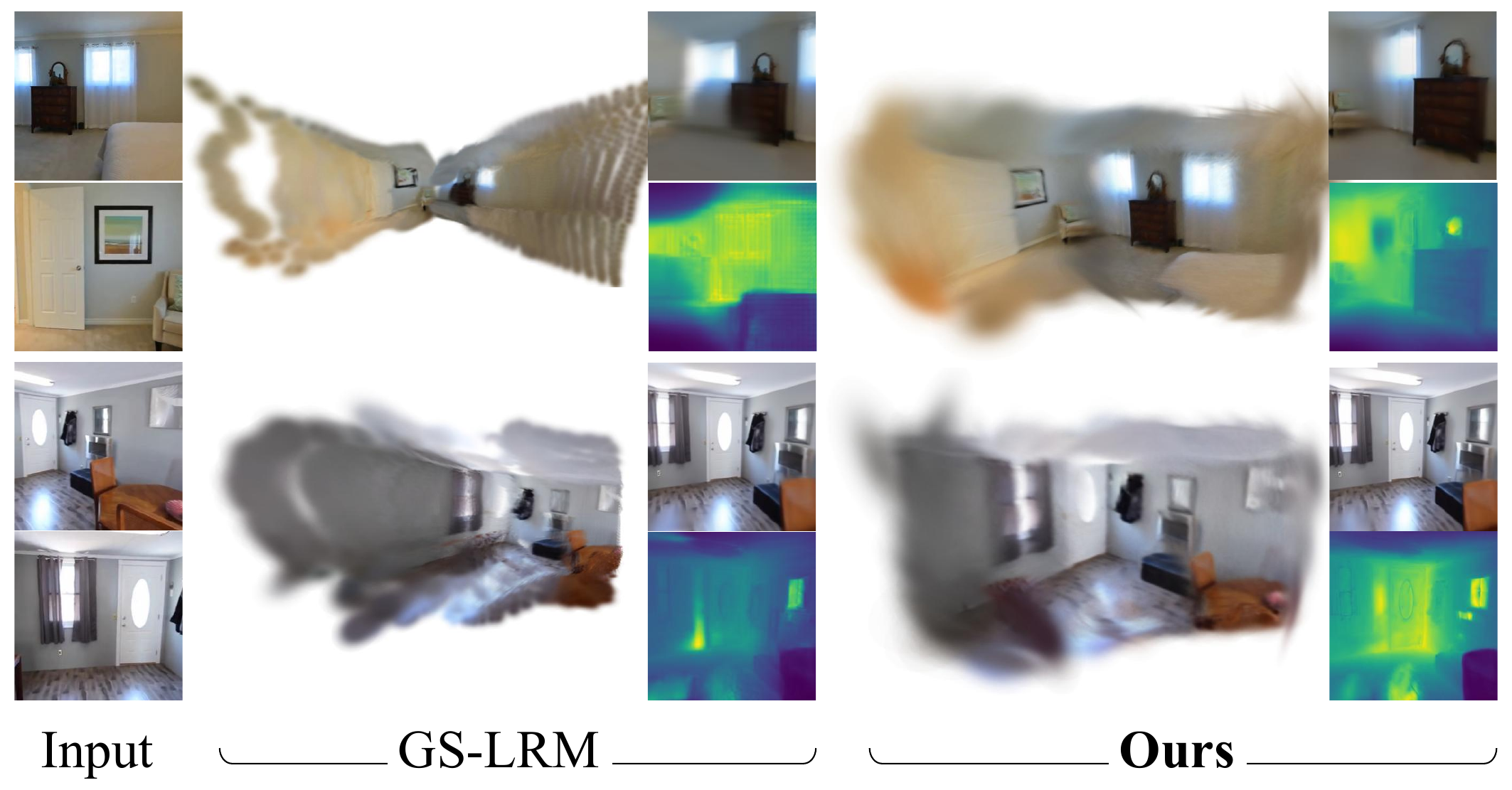} \\
    \vspace{-1em}
    \caption{\textbf{Qualitative results on RE10K~\cite{zhou2018stereo}.} Compared to the pixel-aligned Gaussian prediction of GS-LRM~\cite{zhang2025gs}, our formulation, which is based on direct XYZ prediction and decoupling from pixel rays, produces noticeably cleaner geometry with fewer spiky artifacts.}
    \label{fig:re10k-qual}
    \vspace{-0.5em}
\end{figure}

\parahead{RE10K} We first compare our model with baselines on RE10K~\cite{zhou2018stereo} with two input views.
We evaluate the base model with 1024 Gaussian tokens, the finetuned model with 4096 Gaussian tokens, and the finetuned model with token-tuning.
The results are shown in~\cref{tab:sota_re10k}.
Compared to baselines, the base model achieves on-par performance with $50\%$ of the Gaussians, and the finetuned model wins by a clear margin with $2\times$ of the Gaussians, which showcases our flexible control over the number of Gaussians. Token-tuning effectively improves the reconstruction quality without suffering from overfitting, even with only two training views.
Qualitative comparisons with the GS-LRM~\cite{zhang2025gs} baseline are shown in \cref{fig:re10k-qual}, where one can clearly see the spike artifacts from the 3DGS visualizations.

\parahead{DL3DV} We then evaluate our model on DL3DV~\cite{ling2024dl3dv} with varying number of input views. We fix our model to the finetuned model with 4096 Gaussian tokens, and focus on the generalization to other context lengths. The results are shown in~\cref{tab:dl3dv_compare}. On the 4-view setting which our model is trained, we achieve comparable performance with the baselines with half of the Gaussians. When generalizing to the 2-view and 6-view settings with context extension, our model can still stay competitive with the baselines with no additional finetuning. Notabaly, on the 6-view setting, our model together with TT outperforms the baseline with a clear margin with 72\% fewer Gaussians.

\subsection{View Extrapolation}
\begin{figure}[t]
    \centering
    \includegraphics[width=\linewidth]{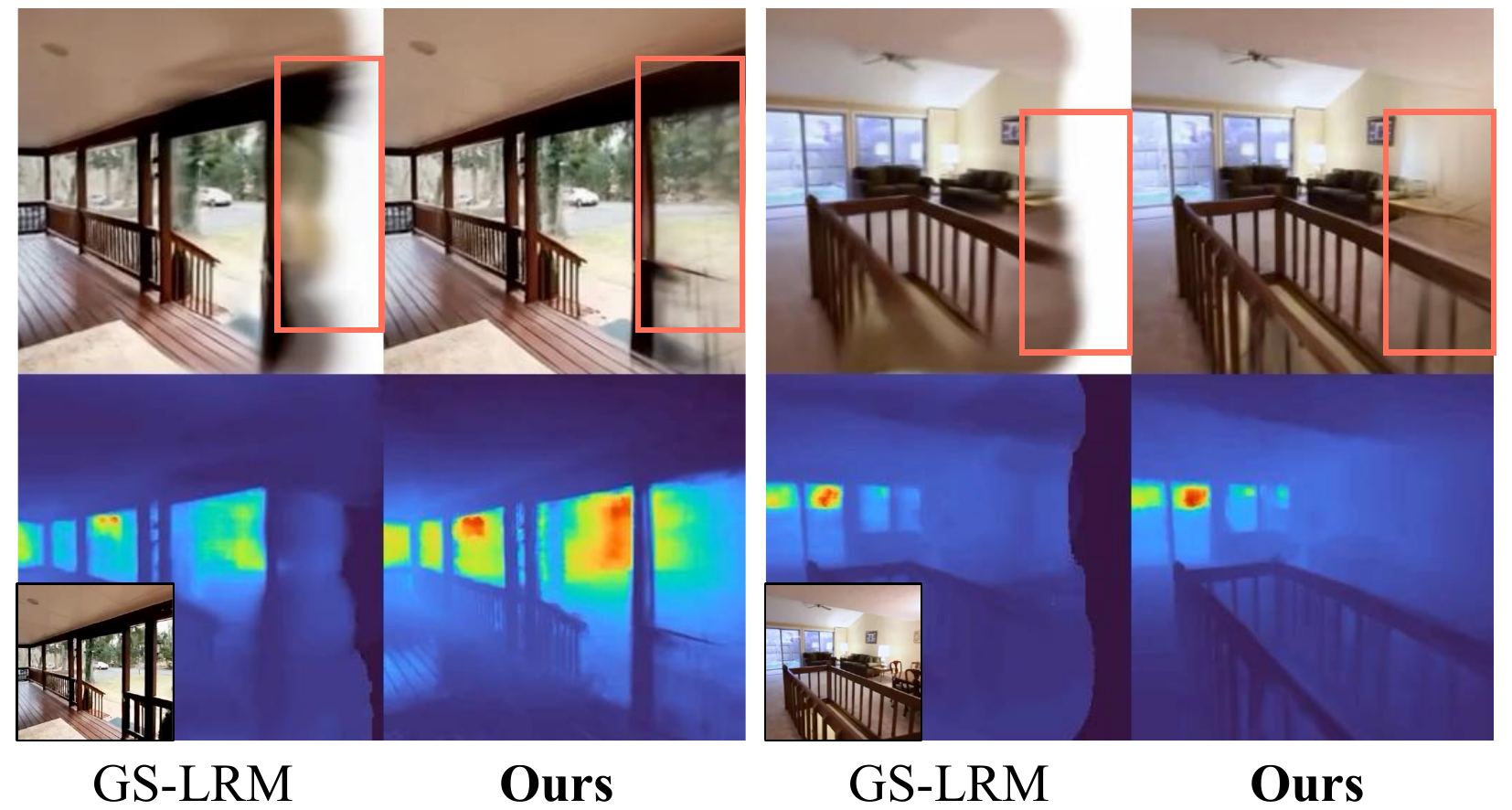}
    \vspace{-2em}
    \caption{\textbf{View Extrapolation}. Both GS-LRM and our model have been finetuned for view extrapolation. The lower-left inset figures show the ground-truth images.}
    \label{fig:view-extrapolation}
    \vspace{-1em}
\end{figure}

We further evaluate our model on a view extrapolation benchmark from~\citep{shi2025revisiting} that samples target views beyond the context window to test the reconstructed geometry on the boundaries that are not observed by the inputs. The results are shown in~\cref{tab:extrapolation}. Without requiring any pointmap or depth annotation, our model outperforms the baselines with a clear margin, and matches the performance of the baselines with pointmap supervised finetuning. 
Additionally, we finetune both GS-LRM and our model specifically for the view extrapolation task (\ie, with target view sampled beyond the context window to test its extrapolation capability), with the visibility loss disabled for ours. The results are shown in~\cref{fig:view-extrapolation}. Since our Gaussian prediction is decoupled from the input rays, our model is able to reconstruct more complete geometry beyond the input cameras, which reinforces the findings from object-level experiments in~\cref{fig:objaverse}.

\subsection{Reconstruction with Camera noise}
\begin{figure}[t]
    \centering
    \includegraphics[width=\columnwidth]{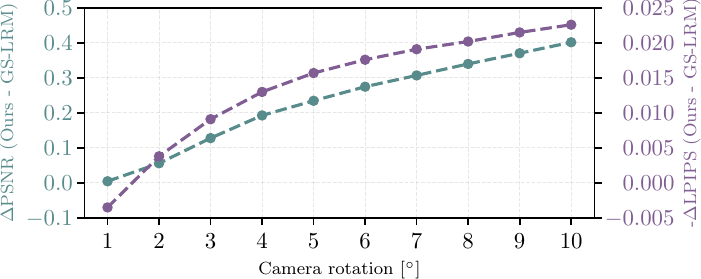}
    \vspace{-2em}
    \caption{\textbf{Reconstruction under camera noise on 2-view RE10K}. We add camera pose noise of magnitude 1–10 degrees to the non-reference view. We show the difference to GS-LRM~\cite{zhang2025gs} in terms of PSNR and LPIPS. Note, that we visualize $-\Delta\text{LPIPS}$ so \textbf{higher values indicate better performance} for both metrics.}
    \label{fig:camera-noise}
    \vspace{-1em}
\end{figure}

We evaluate the robustness of our model to noisy camera poses on RE10K with two input views. To this end, we inject random rotations to the second camera pose by sampling a random axis and applying a rotation with a magnitude of up to 10 degrees. We plot the gap between our model and GS-LRM in terms of PSNR and LPIPS across different noise magnitudes in~\cref{fig:camera-noise}. By decoupling the Gaussian prediction from the input rays, our model is less susceptible to the noise in the input camera poses, and the gap between our model and GS-LRM becomes larger as the noise magnitude increases.

\subsection{Dynamic Reconstruction}
\begin{figure}[t]
    \centering
    \includegraphics[width=\columnwidth]{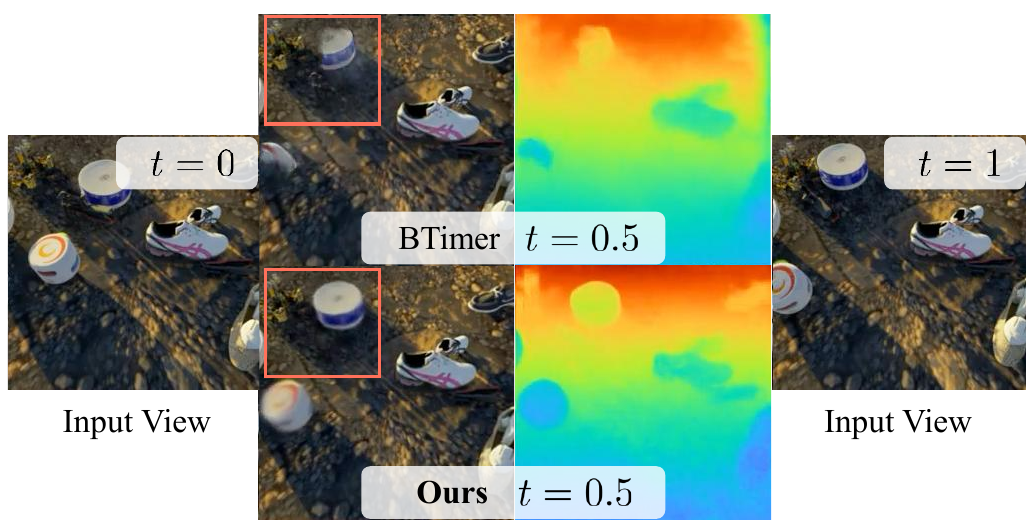}
    \vspace{-2em}
    \caption{\textbf{Dynamic reconstruction on Kubric~\cite{greff2022kubric}}. We show in-between time stamps to examine temporal consistency.}
    \label{fig:kubric}
    \vspace{-1em}
\end{figure}

\begin{figure}[t]
    \centering
    \includegraphics[width=\linewidth]{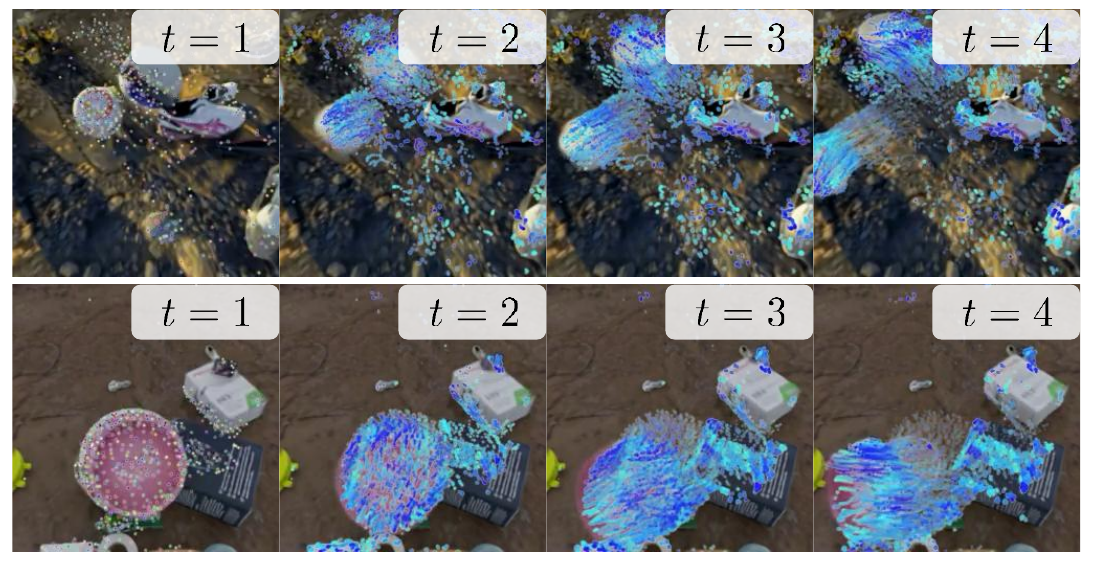}
    \vspace{-2em}
    \caption{\textbf{Emergent scene flow}. We visualize the trajectories of each dynamic Gaussian across time.}
    \label{fig:flow}
    \vspace{-1em}
\end{figure}

We evaluate the dynamic reconstruction capability of our model on Kubric~\citep{greff2022kubric} with 4 input views and resolution $256\times256$. We first train a 4-view base model on both RE10K and DL3DV and then finetune on Kubric with 256 additional dynamic Gaussian tokens for 50K iterations. The evaluation is performed on 16 held-out scenes. A comparison with BTimer~\citep{liang2024feed} is shown in~\cref{fig:kubric}. As the pixel-aligned representation in BTimer prevents it from modeling the motion of Gaussian centers, the model compensates by instead varying the Gaussian opacity across time. As a result, dynamic objects tend to disappear and reappear from one input frame to the next. In contrast, the 3D coordinates that our model predicts are continuous in time and hence better reconstruct object motion. This is especially prominent when target timestamps are between input timestamps. In addition, by conditioning the target time embedding on dynamic Gaussian tokens instead of on input pixels, our model can provide a consistent tracking over time. We visualize the scene flow between different time stamps in~\cref{fig:flow}. A quantitative comparison is shown in~\cref{tab:kubric} where our method outperforms BTimer in terms of PSNR.

\subsection{Ablation Study}

\begin{figure}[t]
    \centering
    \includegraphics[width=0.25\columnwidth]{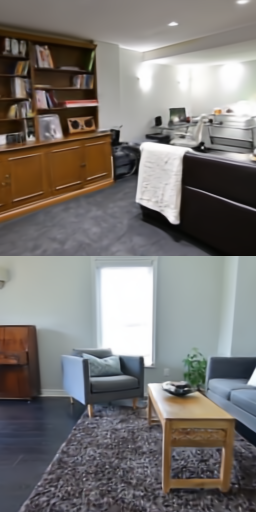}
    \hfill
    \includegraphics[width=0.25\columnwidth]{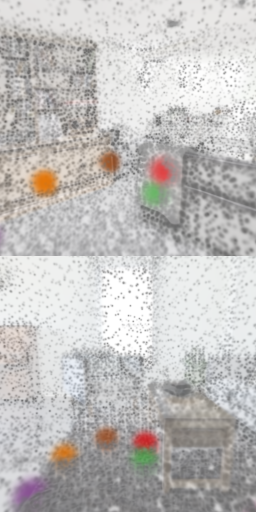}
    \hfill
    \includegraphics[width=0.25\columnwidth]{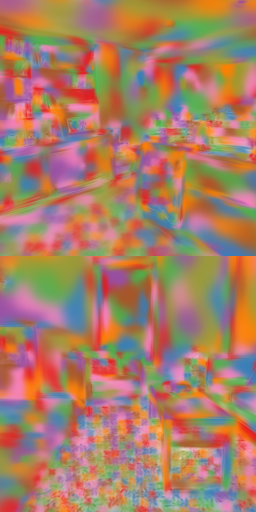}
    \caption{\textbf{Token assignments}. \textbf{Left}: rendering. \textbf{Center}: GS from 5 random tokens highlighted with per-token colors to show their location across scenes. Others set to gray; overlaid with GT image. \textbf{Right}: GS from each token assigned a consistent color.}
    \label{fig:token-correspondences}
    \vspace{-1em}
\end{figure}

\parahead{Visualization of spatial token correspondence}
In~\cref{fig:token-correspondences} we visualize the Gaussians resulting from specific tokens across two scenes and scene decomposition across splats from different tokens.
We find that Gaussians from a single token stay close to each other within a scene and remain in a similar 3d location across scenes -- a phenomenon similar to slot specialization reported in~\cite{carion2020end}.
Looking at the decomposition of the entire image, one can see that tokens can cover regions of uneven areas, with more tokens (and Gaussians) allocated to regions with higher frequency details.

\begin{figure}[t]
    \centering
    \includegraphics[width=\linewidth]{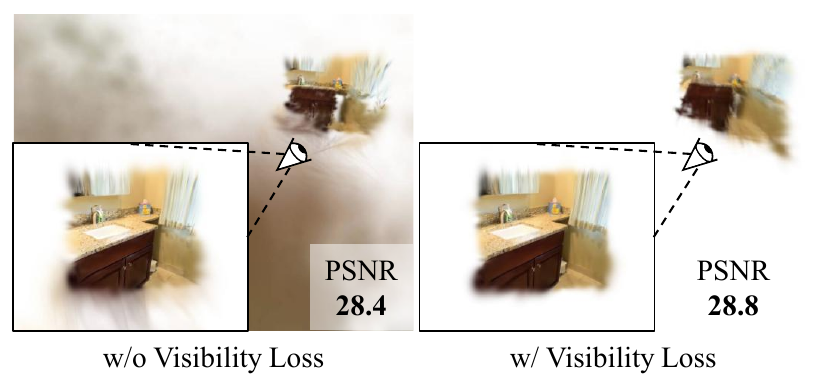}
    \vspace{-2em}
    \caption{\textbf{Effect of the visibility loss}. Regularizing the model removes floaters from unobserved parts of the scene and improves the final PSNR on novel views.}
    \label{fig:visibility-ablation}
    \vspace{-1em}
\end{figure}

\parahead{Visibility Loss}
We evaluate the importance of visibility loss (described in~\cref{sec:gaussian_mean_regression}) by training and evaluating two 4-view models on RE10K, with and without this regularizer.
The visibility loss term prevents degraded geometry, shown in~\cref{fig:visibility-ablation}, and leads to 0.4 PSNR improvement.

\begin{figure}[t]
    \centering
    \includegraphics[width=0.9\linewidth]{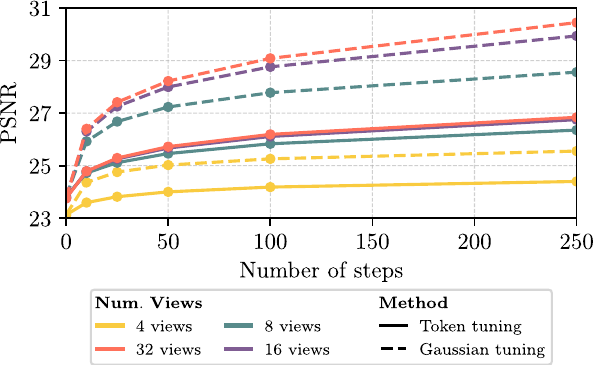} \\
    \vspace{-0.5em}
    \caption{\textbf{Test-time scaling}: PSNR when scaling \#input views vs \#gradient steps in two variants: optimizing Gaussian tokens (solid) or Gaussian parameters directly (dashed).}
    \label{fig:tt-scaling-plot}
    \vspace{-1em}
\end{figure}

\begin{figure}[t]
    \centering
    \includegraphics[width=\linewidth]{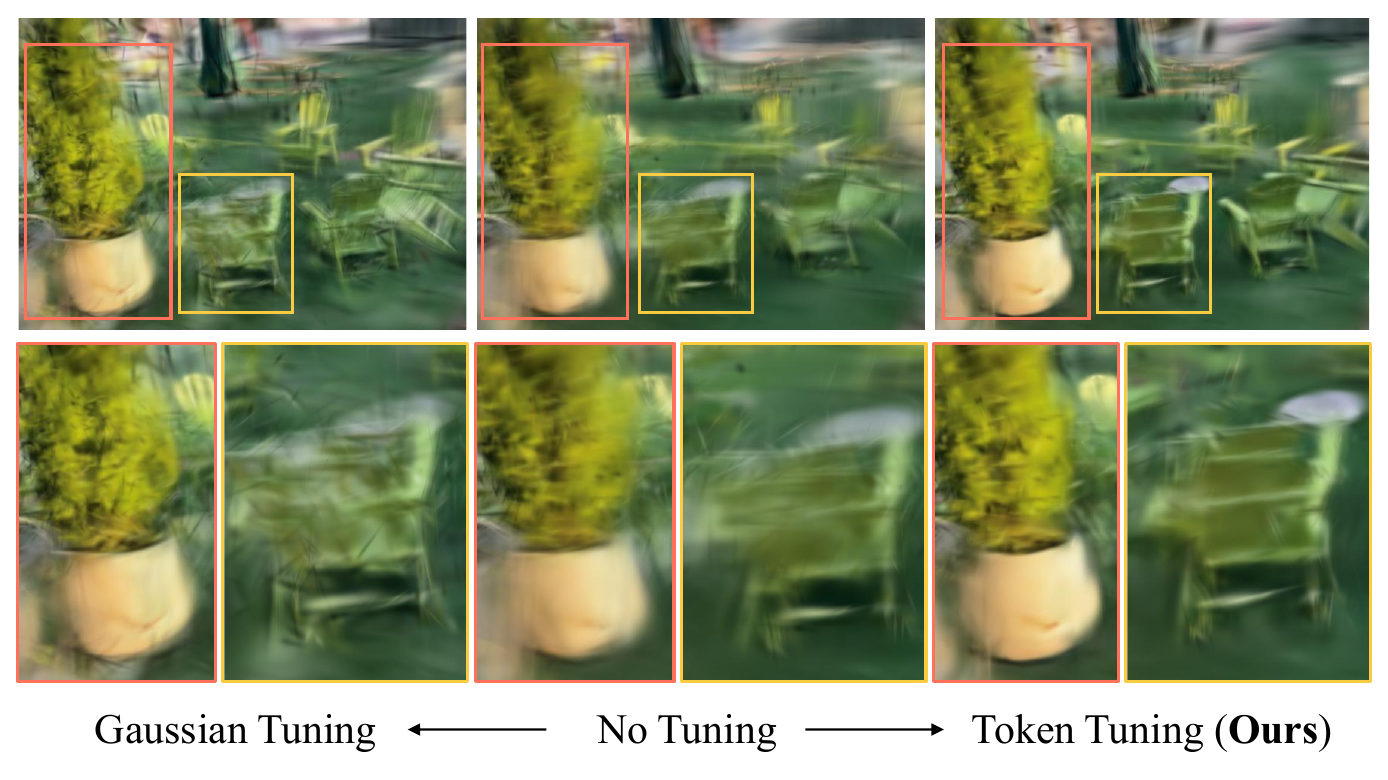}
    \vspace{-2em}
    \caption{\textbf{Center}: renders of a feed-forward reconstruction from 4 views, completely novel viewpoint. \textbf{Left}: Gaussian Tuning degrades scene geometry, despite quantitative advantage on close-by views. \textbf{Right}: Our Token Tuning (TT)  improves scene geometry with sharper renderings. }
    \label{fig:ttt-figure}
    \vspace{-1em}
\end{figure}

\begin{figure}[t]
    \centering
    \includegraphics[width=0.9\linewidth]{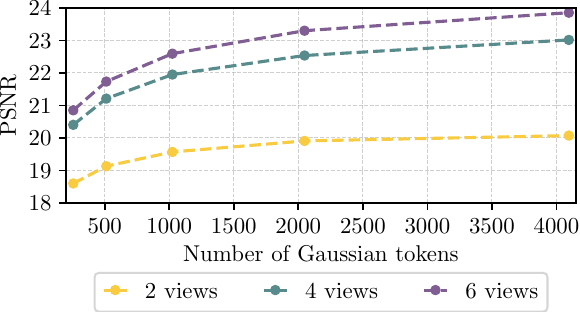}
    \vspace{-0.5em}
    \caption{\textbf{PSNR vs \# Gaussian tokens.} We study how reconstruction quality changes with different numbers of Gaussian tokens given fixed number of input views.}
    \label{fig:psnr_vs_tokens}
    \vspace{-1.5em}
\end{figure}

\parahead{Test-time scaling}
Our model can scale at test time on two axes: the number of input views and the number of token-tuning steps (see~\cref{sec:test_time_training}).
In~\cref{fig:tt-scaling-plot} we compare both strategies on DL3DV~\cite{ling2024dl3dv}. The setting mirrors~\cref{tab:dl3dv_compare}, except that we now sweep over a larger number of input views, uniformly spaced between the first and the last one. We use a 4-view base model and evaluate against fixed 50 target views, which \textbf{include} the input views.

Context extension: our method benefits from up to $4 \times$ more inputs than it was trained for, after which the performance saturates. Token-tuning: even just 10 step optimization of Gaussian tokens leads to significantly improved PSNR, a trend continuing up to 250 steps but slowly saturating. CE and TT can be combined for additional benefit.

We also compare against a simple baseline that directly optimizes Gaussian parameters instead of token embeddings. This baseline achieves  higher PSNR and faster convergence, particularly when many input views are available and the reconstruction problem is less ill-posed. However, as shown in \cref{fig:ttt-figure} , extensive Gaussian-parameter tuning degrades geometry, while token tuning improves it—reflecting the stronger learned prior encoded in the tokens. A deeper exploration of joint optimization strategies (token and Gaussian tuning) is left for future work.
    
\parahead{Effect of number of tokens}
We evaluate the effect of the number of Gaussian tokens on the reconstruction quality on DL3DV. We finetune the 1024-token 4-view base model with number of tokens ranging from 256 to 4096 and number of views ranging from 2 to 6. As shown in~\cref{fig:psnr_vs_tokens}, our approach offers independent control over both the number of Gaussians and input views, providing two-dimensional scalability that pixel-aligned methods lack. This flexibility enables us to adjust the trade-off between compression efficiency and reconstruction quality: fewer Gaussians yield compact representations while more Gaussians achieve higher fidelity, all without changing the number of input views.

\section{Conclusion}
We introduced \MethodName, a feed-forward 3DGS framework that directly regresses the 3D coordinates of the Gaussian centroids and uses learnable Gaussian tokens to decouple the reconstruction from input pixels. This design improves robustness, enhances geometric fidelity, supports dynamic-scene modeling, and enables efficient test-time scaling through context extension and token tuning. 

\parahead{Limitations}
Our method inherits limitations common to feed-forward approaches: it can struggle with large-scale environments and fine-grained geometric detail. Test-time token tuning is also relatively costly, typically requiring several dozen optimization steps.
However, token tuning preserves strong learned priors, suggesting that hybrid strategies combining both approaches present a promising direction for future research.

\appendix

\section{More Implementation Details}
\parahead{Gaussian Token Details} We use a patch size 8 for the decoder, so each Gaussian token decodes 64 Gaussians. We initialize Gaussian tokens with a standard deviation of 0.01. When increasing the number of Gaussian tokens or adding dynamic Gaussian tokens, we initialize the new tokens to existing tokens and apply a small random perturbation with standard deviation 0.01. To make sure most Gaussians are visible at initialization, we initialize the linear output projection layer with a small standard deviation of 2e-3 and initialize the Layer Scale layer in the decoder with 1e-5.

\parahead{Training Details} We apply a gradient clipping with a threshold of 1.0 to the gradients of the model. Weight decay of 0.05 is applied except for the bias terms and the Layer Normalization layers. A random flip augmentation is applied to the images and camera poses during training. The view sampling strategy follows DepthSplat~\cite{xu2025depthsplat}. We rescale the camera translations of RE10K, DL3DV, and Kubric by 0.25, 0.15, and 0.1 respectively to make the mean depths roughly 1. The 150K-iteration base model training takes 36 hours on 8 NVIDIA A100 GPUs and the 10K-iteration finetuning usually takes another 4 hours depending on the number of Gaussian tokens and the resolution.

\parahead{Baseline Details} For comparison, we reproduced GS-LRM~\cite{zhang2025gs} and BTimer~\cite{liang2024feed} following the training details provided in the original papers. We directly compare with the numbers provided in the original papers when available.

{
    \small
    \bibliographystyle{ieeenat_fullname}
    \bibliography{main}
}

\end{document}